\documentclass[sigconf]{acmart}

\usepackage{amsmath}
\usepackage{xcolor}
\usepackage{float}
\usepackage{multirow}
\usepackage{multicol}

\usepackage{caption}
\usepackage{subcaption}

\AtBeginDocument{%
  \providecommand\BibTeX{{%
    \normalfont B\kern-0.5em{\scshape i\kern-0.25em b}\kern-0.8em\TeX}}}

\newcommand\independent{\protect\mathpalette{\protect\independenT}{\perp}}
\def\independenT#1#2{\mathrel{\rlap{$#1#2$}\mkern2mu{#1#2}}}

\setcopyright{acmcopyright}
\copyrightyear{2022} 
\acmYear{2022} 
\setcopyright{acmcopyright}\acmConference[KDD '22]{Proceedings of the 28th ACM SIGKDD Conference on Knowledge Discovery and Data Mining}{August 14--18, 2022}{Washington, DC, USA}
\acmBooktitle{Proceedings of the 28th ACM SIGKDD Conference on Knowledge Discovery and Data Mining (KDD '22), August 14--18, 2022, Washington, DC, USA}
\acmPrice{15.00}
\acmDOI{10.1145/3534678.3539198}
\acmISBN{978-1-4503-9385-0/22/08}

\begin{document}

\title{DESCN: Deep Entire Space Cross Networks for Individual Treatment Effect Estimation}

\author{Kailiang Zhong}
\authornote{Both authors contributed equally to this work.}
\email{brice.zkl@alibaba-inc.com}

\author{Fengtong Xiao}
\authornotemark[1]
\email{fengtong.xiao@alibaba-inc.com}
\affiliation{%
  \institution{Alibaba Group, China, Singapore}
  \country{}
}

\author{Yan Ren}
\email{william.ry@lazada.com}
\affiliation{%
  \institution{Alibaba Group}
  \country{China}
}

\author{Yaorong Liang}
\email{yaorong.lyr@lazada.com}
\affiliation{%
  \institution{Alibaba Group}
  \country{China}
}

\author{Wenqing Yao}
\email{wenqing.ywq@alibaba-inc.com}
\affiliation{%
  \institution{Alibaba Group}
  \country{China}
}

\author{Xiaofeng Yang}
\email{xiaofeng.yang@alibaba-inc.com}
\affiliation{%
  \institution{Alibaba Group}
  \country{Singapore}
}

\author{Ling Cen}
\email{ling.cen@lazada.com}
\affiliation{%
  \institution{Alibaba Group}
  \country{Singapore}
}

\renewcommand{\shortauthors}{Kailiang Zhong et al.}

\begin{abstract}
Causal Inference has wide applications in various areas such as E-commerce and precision medicine, and its performance heavily relies on the accurate estimation of the Individual Treatment Effect (ITE). Conventionally, ITE is predicted by modeling the treated and control response functions separately in their individual sample spaces. However, such an approach usually encounters two issues in practice, i.e. divergent distribution between treated and control groups due to \textit{treatment bias}, and significant \textit{sample imbalance} of their population sizes. This paper proposes Deep Entire Space Cross Networks (DESCN) to model treatment effects from an end-to-end perspective. DESCN captures the integrated information of the treatment propensity, the response, and the hidden treatment effect through a cross network in a multi-task learning manner. Our method jointly learns the treatment and response functions in the entire sample space to avoid treatment bias and employs an intermediate pseudo treatment effect prediction network to relieve sample imbalance. Extensive experiments are conducted on a synthetic dataset and a large-scaled production dataset 
from the E-commerce voucher distribution business. The results indicate that DESCN can successfully enhance the accuracy of ITE estimation and improve the uplift ranking performance. A sample of the production dataset and the source code are released to facilitate future research in the community, which is, to the best of our knowledge, the first large-scale public biased treatment dataset for causal inference\footnote{https://github.com/kailiang-zhong/DESCN}.
\end{abstract}



\maketitle

\section{INTRODUCTION}
Individual-level causal inference is a predictive analytics technique to estimate the Individual Treatment Effect (ITE)\footnote{ITE is sometimes also referred as Conditional Average Treatment Effect (CATE) at the individual level.} of a single or multiple treatments. This technique has wide applications such as identifying the most effective medication to patients \cite{Jaskowski2012}, and optimizing cross-selling for personalized insurance products \cite{Guelman2015a}. It is also popular in the E-commerce domain as a profit-driven business like voucher distribution and targeted advertising. In this paper, we focus on the ITE estimation task where only a single treatment (i.e., treated or non-treated\footnote{Non-treated group is also known as control group.}) exists.

One of the fundamental challenges in the treatment effect modeling (a.k.a uplift modeling) is the existence of \textit{counterfactual outcomes}, that is, we could not observe both treated and control responses in the exact same context for an individual. As a result, the direct measurement of treatment effect from an individual is not observable. A common solution to this challenge is to build models on both response functions separately to derive the counterfactual prediction.

Besides, there are two additional significant issues encountered in practice. First, the underlying distributions for the treated and control groups could be divergent due to the existence of confounding factors. We refer to this as \textit{treatment bias} issue. Confounding factors exist when the treated and control groups are selected not by random, but by some specific factors. Take the E-commerce voucher distribution scenario as an example, a certain discount offer (a.k.a voucher) could only be provided to inactive users with the aim to improve the customer retention rate. The active users, nevertheless, are not given vouchers for cost-saving purposes. As a result, the users in the control group tend to be more active than those in the treated group. It makes a model difficult to learn an unbiased representation between those two groups through loss functions for ITE estimation, like Counter-factual Regression (CFR)\cite{Shalit2017}. 

Second, the sample size between the treated and control group could vary significantly in practice due to particular treatment strategies, which leads to the \textit{sample imbalance} problem. In E-commerce domain, free membership (i.e. treatment) could be given to the majority group of users to promote new products. This makes the sample size of the treated group much larger than the control group. By contrast, in a voucher distribution scenario, it is common that vouchers are only distributed to the users who are promotion-sensitive (i.e. users who will only purchase when a promotion is given), resulting in comparatively small treated group size. Such an imbalanced data could make the model very difficult to learn an accurate estimation of the treatment effects and need extra calibration efforts.

To tackle the above issues, we propose a multi-task cross network to capture the factors of the treatment propensity, the true responses, and the pseudo treatment effect in an integrated framework named Deep Entire Space Cross Networks (DESCN). We introduce an Entire Space Network (ESN) to jointly learn the treatment and response functions in the entire sample space. Instead of separately modeling the Treated Response (TR) and Control Response (CR) in their individual sample space, ESN applies a propensity network to learn the treatment propensity and then connect with TR and CR to derive Entire Space Treated Response (ESTR) and Entire Space Control Response (ESCR). In this manner, the model could be trained on ESTR and ESCR directly where both response functions could leverage the entire samples to address the \textit{treatment bias} issue. Furthermore, it introduces an additional novel cross network (X-network) on top of the shared structure to learn the hidden pseudo treatment effect, which acts as an intermediate variable to connect with the two true response functions (TR and CR) to simulate the counterfactual responses. In this way, we could learn an integrated representation that contains all the responses and the treatment effect information to alleviate the \textit{sample imbalance} issue. 

The main contributions of this paper are as follows:
\begin{itemize}
	\item An end-to-end multi-task cross network, DESCN, is proposed which captures the relationships between the treatment propensity, the true responses, and the pseudo treatment effect in an integrated manner to alleviate \textit{treatment bias} and \textit{sample imbalance} issues simultaneously. Extensive experiments conducted on a synthetic dataset and a large-scale production dataset indicate that DESCN outperforms the baseline models in both ITE estimation accuracy and uplift ranking performance by over +4\% to +7\%.
	\item An Entire Space Network (ESN) for modeling the joint distribution between treatment and response functions in the entire sample space. In this way, we could derive the counterfactual information for both treated and control groups in an integrated manner to leverage from entire samples and address the \textit{treatment bias} issue. Actually, the ESN implies an Inverse Probability Weighting(IPW), which we discuss in detail in section 5.2. Note that the ESN is not limited to the DESCN model but could also be applied to other existing individual response functions estimation based uplift models.
	\item A large-scale production dataset on voucher distribution was collected from the E-commerce platform Lazada. We specifically design the experiment to generate strong treatment bias in the training set but use randomized treatment in the testing set in order to better evaluate the model performance. To the best of our knowledge, this is the first industrial production dataset with both biased and randomized treatments in training and testing set simultaneously, we hope this could help facilitate future research in causal inference.
\end{itemize}

\section{RELATED WORK}
Meta learning \cite{Zhao2019,Zhao2020,Gutierrez2016} is a popular framework to estimate ITE, which uses any machine learning estimators as base learners, and identifies the treatment effect by comparing the estimated responses returned by the base models. S-learner and T-learner are two commonly adopted meta learning paradigms. S-learner is trained over the entire space combining both treated and control samples, with the treatment indicator as an additive input feature. T-learner uses two models built individually on the treatment and control sample spaces. In both S-learner and T-learner, when the population size of the two groups is unbalanced, the performance between the corresponding base models could be inconsistent and hurt ITE estimation performance. To overcome this problem, X-learner \cite{Kunzel2019} is proposed, which adopts information learned from the control group to give a better estimation of the treated group and vice versa. The first step of X-learner is to learn two T-learner like models separately. Then, it calculates the difference between the observed outcome and the estimated outcome from both treated and control groups as the imputed treatment effects, which are further used to train a pair of ITE estimators. Finally, the overall ITE is calculated as the weighted average of the two estimators. As with other meta-learning methods, the performance of the base model still has a strong cascading effect in X-learner.

In meta learning, linear methods (e.g., Least Absolute Shrinkage and Selection Operator (LASSO) \cite{Tibshirani1996}) may not perform well if the treatment indicator feature is less important or even ignored in modeling. Moreover, the difference of sample size between the treated and control group will significantly influence the training loss contribution in the linear model, which means it can't handle \textit{sample imbalance} issue well. Tree-base methods like Bayesian Additive Regression Trees (BART) \cite{Chipman2010} and Causal Forest(CF) \cite{Wager2018} can alleviate the sample imbalanced problem to a certain degree. BART is employed as a base model to estimate heterogeneous treatment effects in the S-learner, which is able to deal with high-dimensional predictors, yields coherent uncertainty intervals as well as handles continuous treatment variables and missing data for the outcome variable. In \cite{Rzepakowski2012}, a classification tree algorithm is designed with the splitting criteria minimizing the heterogeneous treatment effects in the sub-trees. A non-parametric Causal Forest algorithm proposed first builds a large number of causal trees with different sub-sampling rates and then estimates heterogeneous treatment effects by taking an average of the outcomes from these individual uplift trees. It has the advantage of exploiting huge-sized data in a large feature space and abstracting inherent heterogeneity in treatment effects.

Recently, representation learning using neural networks becomes the mainstream treatment distribution debiasing methods. Those methods apply networks consisting of both group-dependent layers and shared layers over the entire treated and control groups, and utilize several extended regularisation strategies to address treatment bias inherent in the observation data \cite{Schwab2019a}. Balancing Neural Network (BNN) \cite{Fredrik2016-co} uses Integral Probability Metric (IPM) to minimize the discrepancy between the distributions of treatment and control samples. Treatment-Agnostic Representation Network (TARNet) and Counterfactual Regression (CFR) \cite{Shalit2017} extend BNN into a two-headed structure in the multi-task learning manner, learning two functions based on a shared and balanced feature representation across the treated and control spaces. The similarity-preserved individual treatment effect (SITE) estimation method \cite{Yao2018} improves the CFR by adding a position-dependent deep metric (PPDM) and middle-point distance minimization (MPDM) constraints, which not only balances the distributions of the treated and control population but also preserves the local similarity information. Furthermore, the Perfect Match (PM) method \cite{Schwab2019b} extends the TARNet to multi-treatments scenarios, augmenting the samples within a min-batch and matching their propensity with nearest neighbours. An adapting neural network, DragonNet \cite{Shi2019} is designed for predicting propensity score and conditional outcome in an end-to-end procedure. However, those methods do not address the previously mentioned two issues together.

In recommendation systems, the Entire Space Multi-Task Model (ESMM) \cite{Ma2018EntireSM} has been introduced to solve the sample selection bias problem by exploiting the sequential pattern of user behaviors to model both user clicking and user purchasing in the entire space. In this paper, we apply a similar idea of multi-task learning and entire sample space modeling to causal inference domain. We exploit the structural properties among the treatment propensity, the true response, and the pseudo treatment effect to estimate ITE in the entire sample space to alleviate the \textit{treatment bias} and \textit{sample imbalance} issues.

\section{METHODOLOGY}
This section introduces the DESCN model for ITE estimation. It starts with the problem definition, followed by the illustration of DESCN and its two components: Entire Space Network (ESN) and X-network.

\subsection{Problem Definition}
\label{problem def}
We follow the Neyman-Rubin potential outcome framework \cite{Rubin2005CausalIU} to define the ITE estimation problem. To be specific, assuming we observe samples $\mathcal{D}=\{(y_i, x_i, w_i)\}^n_{i=1}$, with $(Y, X, W) \stackrel{i.i.d.}{\sim} \mathbb{P}$. Here, for each sample $i$, $y_i \in \{0, 1\}$ is a binary treatment outcome of interest; $x_i \in \mathcal{X} \subset \mathbb{R}^d$ is a $d$-dimensional covariate or feature vector; $w_i\in \{0, 1\}$ denotes the binary treatment. Let $y_i(1)$ and $y_i(0)$ denote the potential outcome of individual $i$ when $i$ is being treated ($w_i = 1$) or non-treated ($w_i = 0$). The treatment is assigned according to the propensity score $\pi(x) = P(W=1|X=x) $. 

Further, let $T= \{i: w_i = 1\}$ and $C= \{i: w_i = 0\}$ denotes the set of treated samples and control samples respectively. In this way, $T$ and $C$ represent the treated and control sample space, and we refer to them as sub-sample spaces. The union of $T$ and $C$ represents the whole sample set and referred as the entire sample space.

The major challenge in ITE estimation is that only one potential outcome, either $y_i(0)$ or $y_i(1)$, is observable for each individual $i$ but not both, that is, $y_i = w_i y_i(1) + (1 - w_i) y_i(0)$, where $w_i \in \{0,1\}$. Hence, there is no true uplift value of each sample $y_i(1) -y_i(0)$. Our interest is to estimate the expected individual treatment effect (ITE) with covariate values $X = x$, denoted as $\tau(x)$:
\begin{align}
\begin{aligned}
\tau(x) = \mathbb{E}_\mathbb{P}(Y(1)-Y(0)|x),
\end{aligned}
\end{align}
We consider the problem under three standard assumptions:
\begin{itemize}
    \item Consistency: if individual $i$ is assigned treatment $w_i$, we observe
    a consistent associated potential outcome $y_i=y_i(w_i)$. 
    \item Ignorability: there are no unobserved confounders, such that $Y(1), Y(0) \independent W | X$. 
    \item Overlap: treatment assignment is non-deterministic, i.e. $0 < \pi(x) < 1 \text{, }\forall x\in \mathcal{X}$. 
\end{itemize}
Treated Response (TR) and Control Response (CR) are the outcome of interest under the treated and non-treated (control) groups, which are denoted as $\mu_1(x)$ and $\mu_0(x)$:
\begin{align}
\begin{aligned}
\mu_1(x) &= \mathbb{E}_\mathbb{P}(Y|W=1,X=x),  \\
\mu_0(x) &= \mathbb{E}_\mathbb{P}(Y|W=0,X=x)  
\end{aligned}
\end{align}
In this way, ITE could be written as $\tau(x) = \mu_1(x) - \mu_0(x)$. In order to estimate $\tau$ from observational data, a natural idea is to estimate $\mu_1$ and $\mu_0$ on the treated and control sub-sample space respectively,  and then get $\hat\tau = \hat\mu_1 - \hat\mu_0$\footnote{$\hat{f}$ denotes the estimation of the true function $f$.}. However, such a method has the following two problems:

\begin{itemize}
    \item \textbf{Treatment Bias:} As the treatment is assigned according to the propensity score $\pi$, there is systematic difference between the treated and control group data distributions.
    \item \textbf{Sample Imbalance:} The population size of treated and control subspace could vary significantly. 
\end{itemize}

In order to address those two issues, we propose a novel Entire Space Network (ESN) and a multi-task cross network (X-network). The combination of ESN and X-network gives us Deep Entire Space Cross Networks (DESCN). Details of each network will be introduced in the following sections.

\begin{figure*}
\vspace*{-0.1cm} 
\resizebox{1\linewidth}{!}{
    \begin{subfigure}{.3\textwidth}
      \centering
      \includegraphics[width=1\linewidth]{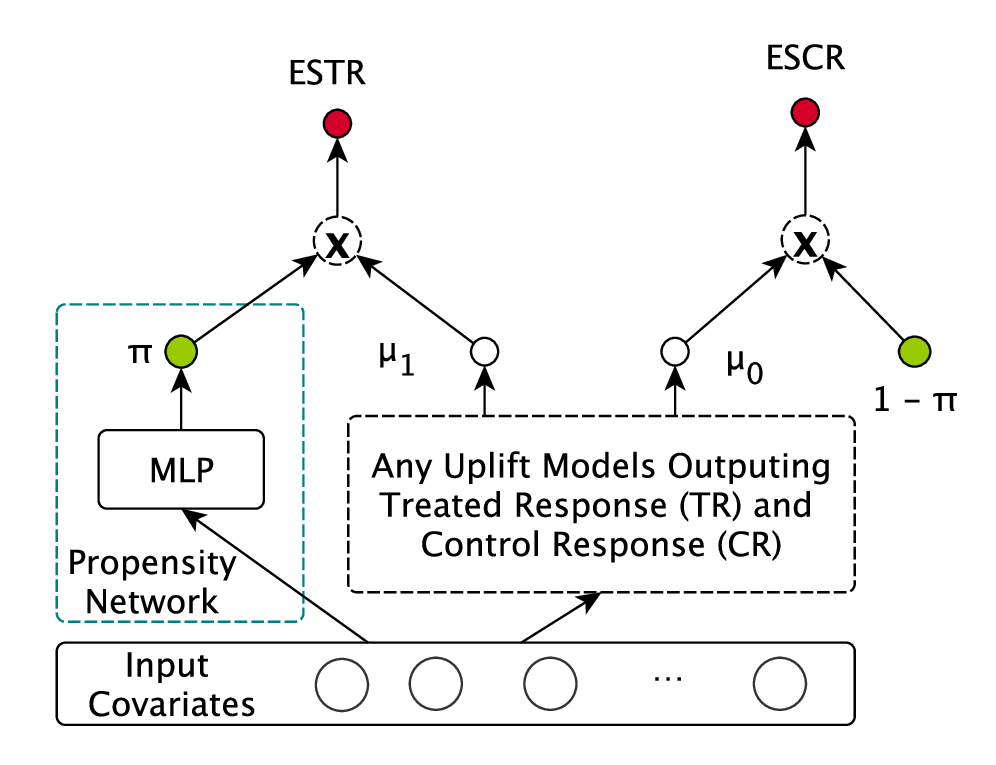}
      \caption{Entire Space Network (ESN)}
      \label{fig:ESN}
    \end{subfigure}%
    \hspace{-0.3cm}
    \begin{subfigure}{.3\textwidth}
      \centering
      \includegraphics[width=1\linewidth]{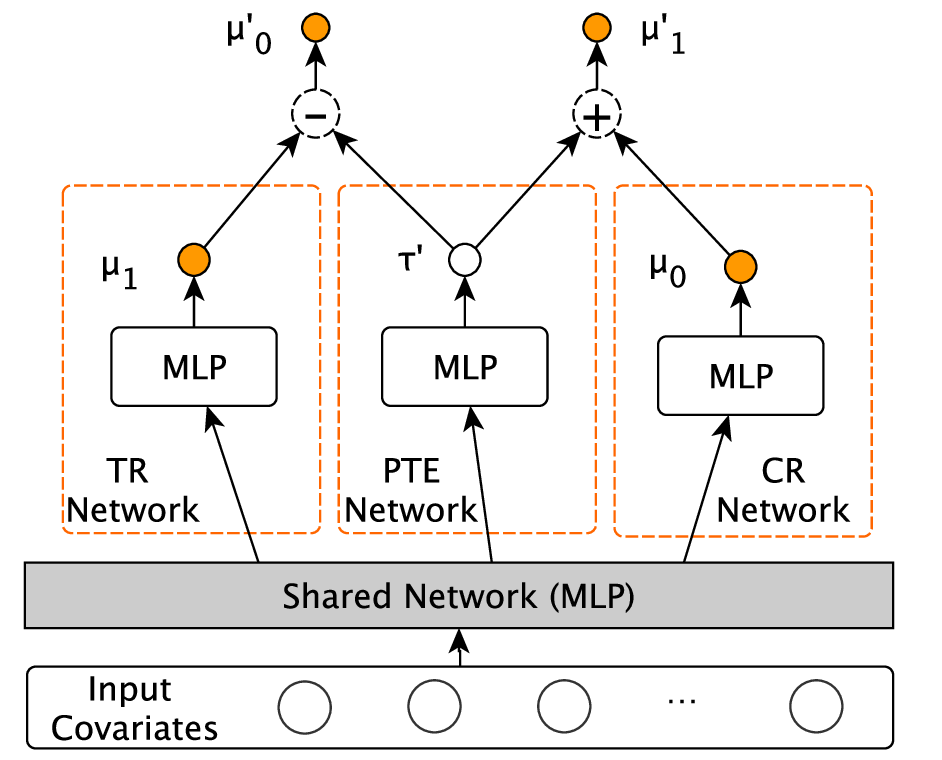}
      \caption{X-network}
      \label{fig:Xnetwork}
    \end{subfigure}%
    \begin{subfigure}{.4\textwidth}
      \centering
      \includegraphics[width=1\linewidth]{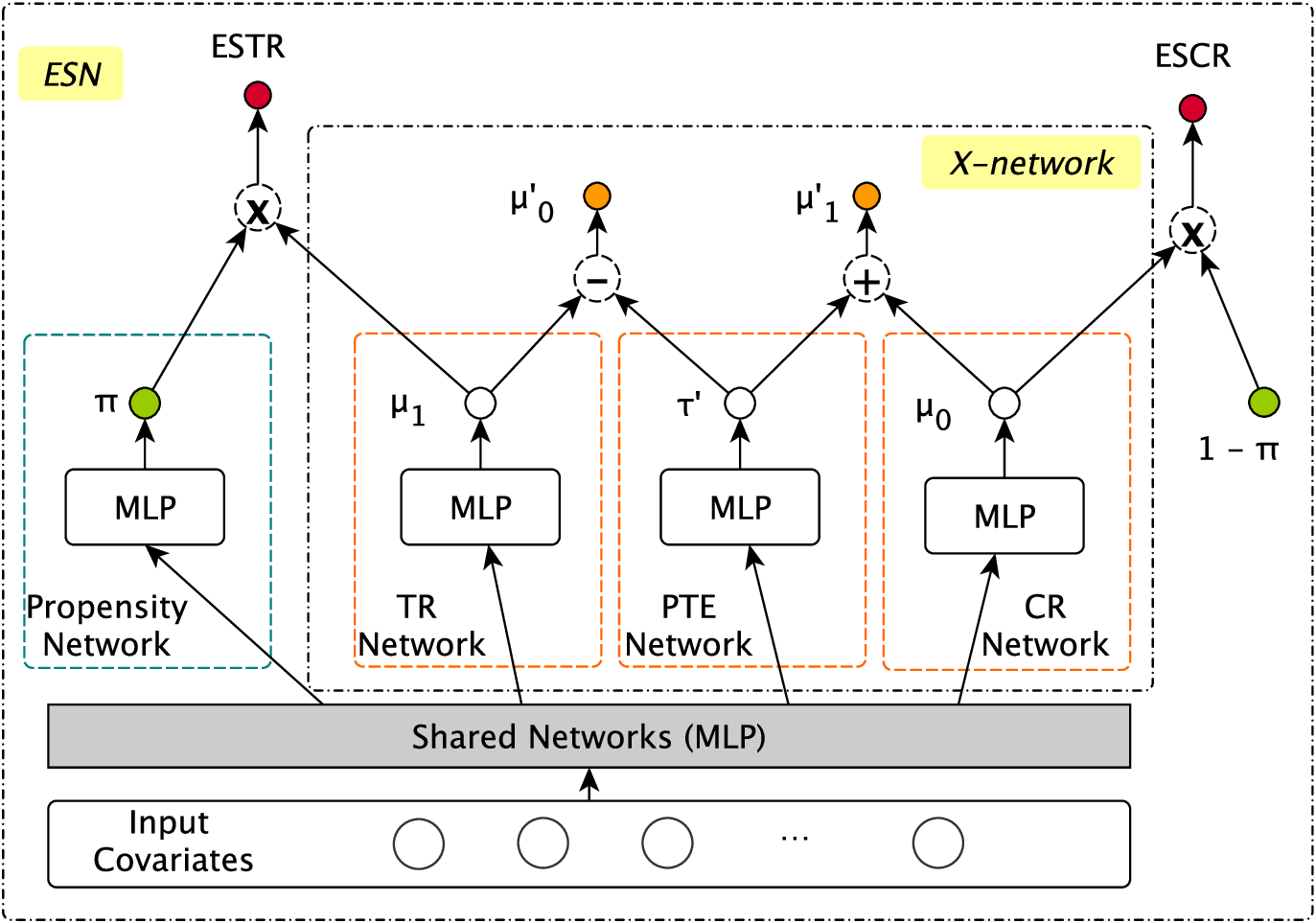}
      \caption{Deep Entire Space Cross Networks (DESCN)}
      \label{fig:DESCN}
    \end{subfigure}}
\setlength{\abovecaptionskip}{-0.01cm}
\setlength{\belowcaptionskip}{-0.5cm}
\vspace{0.1cm}
\caption{Model Architectures of Entire Space Network (ESN), X-network, and DESCN. Note that DESCN is the combination of ESN and X-network. All nodes associated with training loss are highlighted in color.}
\vspace{0.4cm}
\label{fig:main}
\end{figure*}

\subsection{Entire Space Network (ESN)}
\label{Method ESN}
The estimation of ITE requires the estimation of $\mu_1$, $\mu_0$. Inspired by ESMM\cite{Ma2018EntireSM}, we define two associated probabilities: Entire Space Treated Response (ESTR) as $P(Y,W=1|X)$, and the Entire Space Control Response (ESCR) as $P(Y,W=0|X)$. Their probabilities follows Eq.\eqref{ESTR}:
\begin{align}
\label{ESTR}
\begin{aligned}
\underbrace{P(Y, W=1|X)}_{ESTR} &= \underbrace{P(Y|W=1,X)}_{TR} \cdot \underbrace{P(W=1|X)}_{\pi} \\ &= \mu_1 \cdot \pi \\
\vspace{0.5cm}
\underbrace{P(Y, W=0|X)}_{ESCR} &= \underbrace{P(Y|W=0,X)}_{CR} \cdot \underbrace{P(W=0|X)}_{1-\pi} \\ &= \mu_0 \cdot (1-\pi)
\end{aligned}
\end{align}
Note that ESTR, ESCR, and propensity score could be modeled on all samples to get the corresponding unbiased estimated values. Hence, $\hat{\mu_1}$ and $\hat{\mu_0}$ could be derived from them by dividing the estimated ESTR and ESCR by $\pi$ and $1-\pi$ respectively. We could get the estimation of ITE thereafter.

With such structural properties in conditional probabilities, we propose an Entire Space Network (ESN) as shown in Figure \ref{fig:ESN}. The leftmost part of the network is to model the propensity score $\pi(X)$, with the output node $\pi$. On the right are two additional nodes $\mu_1$ and $\mu_0$, representing the modeling of TR and CR respectively. The output of $\mu_1$ node and $\pi$ node are multiplied to get ESTR, while the output of $\mu_0$ is multiplied with 1 - $\pi$ to get ESCR as shown in Eq.\ref{ESTR}. In this way, instead of learning $\mu_0$ and $\mu_1$ in their individual sub-sample spaces, we derive them in integration by modeling the ESTR, ESCR, and $\pi$ in the entire sample space. As a result, a sample from the treated group will not only contribute to the learning of treated response function but also contribute to the learning of control response function, and vice versa. The counterfactual information for both treated and control groups could be derived in the integrated manner from entire samples, which could alleviate the \textit{treatment bias} issue. Note that ESN has to be used with other uplift models that output the estimation of TR and CR, such as TARNet, CFR \cite{Shalit2017}, X-network, etc. A more specific performance comparison on adding ESN to those existing methods is included in Section \ref{ESN Ablation}.

The training loss function of ESN is the weighted sum of the following\footnote{\& denotes the logical AND operation between two binary label $y_i$ and $w_i$.}

\begin{align}
    \label{Loss ESN}
    \begin{aligned}
    L_{\pi} &= \frac{1}{n} \sum_{i}l(w_i, \hat\pi(x_i)) \\
    L_{ESTR} &= \frac{1}{n} \sum_{i}l\big(y_i\&w_i, \hat\mu_1(x_i) \cdot \hat\pi(x_i)\big) \\
    L_{ESCR} &= \frac{1}{n} \sum_{i}l\big(y_i\&(1-w_i), \hat\mu_0(x_i) \cdot (1-\hat\pi(x_i))\big) \\
    \end{aligned}
\end{align}

\vspace{0.6cm}
that is,
\begin{align}
    \begin{aligned}
    L_{ESN} &= \alpha \cdot L_{\pi} + \beta_1 \cdot L_{ESTR} + \beta_0 \cdot L_{ESCR}
    \end{aligned}
    \vspace{-0.3cm}
\end{align}
where $l(\cdot)$ denotes the loss function (i.e. cross entropy loss) and $\alpha$, $\beta_1$, $\beta_0$ are the hyper-parameters denotes the weight of corresponding loss functions.

\begin{table*}[t]
    \centering
    \resizebox{0.8\linewidth}{!}{
    \begin{tabular}{c|c|cccc|cccc}
        \toprule
        \multirow{2}{*}{\centering Dataset} &\multirow{2}{*}{\centering Covariates}& \multicolumn{4}{c|}{Training Data} & \multicolumn{4}{c}{Testing Data} \\
         & & \multirow{2}{*}{Treated} & \multirow{2}{*}{Control} & \multirow{2}{*}{Total} & Positive Outcome & \multirow{2}{*}{Treated} & \multirow{2}{*}{Control} & \multirow{2}{*}{Total} & Positive Outcome \\
         & & & & & (percentage) & & & & (percentage) \\
         \midrule
        Epilesy & 178 & 20.2K &	19.8K &	40k	& 18.4k (45.9\%) & 19.8K & 20.1K & 40k & 18.1k (45.3\%) \\
        Production & 83 & 0.92M & 3.25M & 4.17M	& 83.0k (2.0\%) & 0.47M & 0.43M & 0.91M & 31.9k (3.5\%) \\
        \bottomrule
    \end{tabular}}
    \vspace{0.1cm}
    \caption{Statistics of the synthetic dataset \textit{Epilesy} and real-world Production Dataset.}
    \vspace{-0.6cm}
    \label{table:dataset}
\end{table*}

\vspace{-0.4cm}
\subsection{X-network}
\label{Method Xnetwork}
Inspired by X-learner \cite{Kunzel2019},  to further address the \textit{sample imbalance} issue, we propose a cross network named X-network as shown in Figure \ref{fig:Xnetwork}. Besides the regular networks modeling Treated Response (TR Network) and Control Response (CR Network), we introduce the Pseudo Treatment Effect network (PTE network) to learn the Pseudo Treatment Effect (PTE) ${\tau'}$, that is, the hidden observable effect brought by the treatment. PTE variable ${\tau'}$ acts as an intermediate variable to connect with both TR and CR, which can balance the learning between those two response functions during the training process. 

We add the outputs from the PTE Network and the CR Network to derive the Cross Treated Response $\mu'_{1} := \mu_{0} + \tau'$, which represents the counterfactual results if the individuals of the control group were given treatment. Similarly, we deduct the pseudo treatment effect from TR to derive the Cross Control Response $\mu'_{0} := \mu_{1} - \tau'$, which represents the counterfactual results when the individuals of the treated group did not receive treatment. With these additional nodes, we build the connection between TR and CR through the PTE $\tau'$, and learn a shared representation in the bottom shared network. This helps to learn both TR and CR in a more balanced and constrained manner, especially when one of the treated or control sample size is significantly smaller or larger than the other. Note that we do not force the model to learn a similar representation across treated and control groups through regularization as introduced in CFR\cite{Shalit2017}, instead, we utilize the PTE to connect those two groups. In this manner, we are not smoothing over different units and thus improve the ITE estimation performance. The X-network has 4 losses in the treated or control sub-sample spaces defined as:
\begin{align}
    \label{Loss Xnetwork}
    \begin{aligned}
    L_{TR} &= \frac{1}{|T|} \sum_{i \in T} l(y_i, \hat\mu_1(x_i)), \\
    L_{CR} &= \frac{1}{|C|} \sum_{i \in C} l(y_i, \hat\mu_0(x_i)), \\
    L_{Cross TR} &= \frac{1}{|T|}\sum_{i \in T} l(y_i, \hat\mu'_1(x_i)) \\
        &= \frac{1}{|T|} \sum_{i \in T} l\Big(y_i, \sigma\big(\sigma^{-1}(\hat\mu_0(x_i)) + \sigma^{-1}(\hat\tau'(x_i)\big)\Big), \\
    L_{Cross CR} &= \frac{1}{|C|} \sum_{i \in C} l(y_i, \hat\mu'_0(x_i)) \\
        &= \frac{1}{|C|} \sum_{i \in C} l\Big(y_i, \sigma\big(\sigma^{-1}(\hat\mu_1(x_i)) - \sigma^{-1}(\hat\tau'(x_i)\big)\Big)
    \end{aligned}
\end{align}
where $\sigma$ is the sigmoid function, $L_{TR}$ and $L_{CR}$ represents the loss for Treated Response and Control Response, $L_{CrossTR}$ and $L_{CrossCR}$ represents the loss for Cross Treated Response and Cross Control Response respectively. 

Note that instead of learning the direct values of $\mu_1$ and $\mu_0$, we learn their logit value (i.e. the input of sigmoid function) in TR network and CR network respectively for numerical stability. Specifically, $\hat{\tau}'$ is added(subtracted) on logits of $\hat\mu_0$($\hat\mu_1$) to drive $\hat{\mu_{1}}'$($\hat{\mu_{0}}'$), i.e. $\hat{\mu_{1}}' = \sigma\big(\sigma^{-1}(\hat\mu_0) + \sigma^{-1}(\hat{\tau}')\big)$, and likewise for $\hat{\mu_{0}}'$. Another benefit of such transformation is that we do not need worry about the addition or subtraction getting out of the range $[0,1]$. Besides, when $\hat\mu_0$ and $\hat\mu_1$ is very close to 0 or 1, the $\sigma^{-1}$ function could magnify the uplift signal and make the MLP learning process easier. Empirically, we find that this transformation performs better than directly outputting the true response values.

We can see that our X-network is similar to X-learner in that both try to directly learn the counterfactual treatment effect. In X-learner, ITE is learned based on the results from base learners, and its performance is heavily subject to that of the base models. By contrast, in X-network, ITE is learned together with the base learners in an integrated way.  




\subsection{Model Architecture of DESCN}
The overall model architecture of DESCN is presented in Figure \ref{fig:DESCN}, which is the direct application of Entire Space Network (ESN) on X-network. In this way, both \textit{treatment bias} and \textit{sample imbalance} mentioned in \ref{problem def} could be alleviated. Furthermore, the shared network in DESCN could learn both the propensity score and the control response with pseudo treatment effect simultaneously, which could also help learn a comprehensive representation capturing all this information. 

For the overall training, we have the final loss for DESCN which is the weighted sum of the propensity score loss $L_{\pi}$, the Cross Treated Response $L_{Cross TR}$, and Cross Control Response losses $L_{Cross CR}$:
\begin{align}
    \begin{aligned}
        L_{DESCN} &= L_{ESN} + \gamma_1 \cdot L_{Cross TR} + \gamma_0 \cdot L_{Cross CR} \\
            &= \alpha \cdot L_{\pi} + \beta_1 \cdot L_{ESTR} + \beta_0 \cdot L_{ESCR} \\ 
            &+ \gamma_1 \cdot L_{Cross TR} + \gamma_0 \cdot L_{Cross CR}
    \end{aligned}
\end{align}
Note that we remove $L_{TR}$ and $L_{CR}$ losses in Eq.\eqref{Loss Xnetwork} in section \ref{Method Xnetwork} since TR and CR are now connected with propensity $\pi$ as ESTR and ESCR, which are trained in the entire sample space instead.

\begin{table*}[t]
  \resizebox{1\linewidth}{!}{
  \begin{tabular}{ccccccccccccccc}
    \toprule
    Model & \multicolumn{3}{c}{Epilepsy Dataset} &  & \multicolumn{3}{c}{Production Dataset} &\\
    \cline{2-4}\cline{6-8}
    & $ \sqrt{\epsilon_{PEHE}}$ & Impr & $\epsilon_{ATE}$ & & AUUC & Impr  & $\epsilon_{ATT}$\\
    & mean $\pm$ s.e. & ($CFR_{mmd}$) & mean $\pm$ s.e. & &  mean $\pm$ s.e. & ($CFR_{mmd}$) & mean $\pm$ s.e.\\
    \midrule
X-leaner (NN) & 0.1556 $\pm$ 0.0018 & -15.8\% & 0.0378 $\pm$ 0.0059 & & 0.0234 $\pm$ 0.0035 & -27.9\% & 0.0076 $\pm$ 0.0009 \\
Causal Forest & 0.1519 $\pm$ 0.0042 & -13.0\% & 0.0663 $\pm$ 0.0086 & & 0.0132 $\pm$ 0.0008  & -59.2\% & 0.0123 $\pm$ 0.0003 \\
BART & 0.1387 $\pm$ 0.0004 & -3.2\% & 0.0389 $\pm$ 0.0004 & & 0.0222 $\pm$ 0.0003 & -31.5\% & 0.0312 $\pm$ 0.0001 \\
TARNet & 0.1373 $\pm$ 0.0028 & -2.2\% & 0.0405 $\pm$ 0.0091 & & 0.0309 $\pm$ 0.0021 & -4.6\% & 0.0106 $\pm$ 0.0016 \\
$CFR_{wass}$ & 0.1363 $\pm$ 0.0031 & -1.4\% & 0.0263 $\pm$ 0.0097 & & 0.0261 $\pm$ 0.0002 & -19.4\% & 0.0266 $\pm$ 0.0013 \\
$CFR_{mmd}$ & 0.1344 $\pm$ 0.0027 & 0.0\% & 0.0305 $\pm$ 0.0069 & & 0.0324 $\pm$ 0.0029 & 0.0\% & 0.0258 $\pm$ 0.0015 \\
X-network & 0.1289 $\pm$ 0.0026 & 4.1\% & 0.0245 $\pm$ 0.0044 & & 0.0324 $\pm$ 0.0016 & 0.0\% & 0.0048 $\pm$ 0.0010 \\
\textbf{DESCN} & \textbf{0.1241 $\pm$ 0.0009} & \textbf{7.6\%} & \textbf{0.0058 $\pm$ 0.0014} & & \textbf{0.0340 $\pm$ 0.0006} & \textbf{4.9\%} & \textbf{0.0039 $\pm$ 0.0007} &\\
  \bottomrule
\end{tabular}
}
\caption{Model performance evaluated by $\epsilon_{PEHE}$, $\epsilon_{ATE}$ on the Epilepsy Dataset and by AUUC, $\epsilon_{ATT}$ on the Production Dataset with corresponding mean and standard error. Note that for $\epsilon_{PEHE}$, $\epsilon_{ATE}$ and $\epsilon_{ATT}$, smaller value is better, and for AUUC larger value is better. Best results of all methods are highlighted in boldface.}
\vspace{-0.5cm}
\label{table:main}
\end{table*}

\section{Experiments}
\subsection{Dataset}
To compare the model performance from both ITE estimation accuracy and uplift ranking perspectives, we use a synthetic dataset where the ground truth treatment effect is available and another real-world dataset where the \textit{treatment bias} and \textit{sample imbalance} issues exist. The statistics of those two datasets are summarized in Table \ref{table:dataset}.

The first dataset is a synthetic dataset generated based on the public code from the 2019 American Causal Inference Conference (ACIC) data challenge\footnote{https://sites.google.com/view/acic2019datachallenge/home}. We use the provided Data Generator Processor (DPG) on the covariates collected from the high dimensional Epileptic Seizure Recognition Dataset (\textit{Epilepsy}) \cite{Andrzejak2001} to generate 40k observational samples\footnote{We adopt method that involves complex models and treatment effect heterogeneity, indicated as \textit{Mod 4} in the original source code.}. The advantage of this synthetic dataset is that the ground truth treatment effect is fully known from DPG, thus the \textit{Epilepsy} dataset is ideal to evaluate the model performance on ITE estimation accuracy.

To further evaluate the model performance on uplifting ranking performance, we use another large-scale production dataset from the real voucher distribution business scenario in Lazada, a leading South-East Asia (SEA) E-commerce platform of Alibaba Group. In the real production environment, the treatment assignment is selective due to the operation targeting strategy, and we collect those data as our training set which includes strong treatment bias. We also have a slightly smaller size of users who are not affected by the targeting strategy and the treatment assignment follows the randomized controlled trials (RCT). We use them as our testing dataset. 

By the above setup, the training set contains the biased treated data while the testing set only contains unbiased treated data. In the training set, the treated and control distribution is naturally divergent due to \textit{treatment bias} issue and the sample size between the two groups differ naturally in the real-world scenario as shown in Table \ref{table:dataset}.
\subsection{Compared Methods}
We compare DESCN with the following models which are commonly adopted in ITE estimation to evaluate its performance. All models use all input dense covariates in the two datasets.
\begin{itemize}
	\item \textbf{X-learner (NN)}\cite{Kunzel2019}: We take X-learner as the representative meta-learners. Neural Network (NN) is chosen as the base learner, which has a better non-linear learning advantage and provides a fair comparison with our proposed NN-based model. Two response fitting models, two imputed treatment effects fitting sub-models, and one propensity model are included. All models are trained separately without the shared network parameters. Estimated propensity scores are used for the final ITE output for better performance.
	\item \textbf{BART}\cite{Chipman2010}: Bayesian Additive Regression Trees (BART) is a sum-of-trees model, used to assess the performance of non deep-learning models.
	\item \textbf{Causal Forest}\cite{Wager2018}: Causal Forest is a non-parametric Random Forest based tree model that directly estimates the treatment effect, which is another representative of tree-based uplift models.
	\item \textbf{TARNet}\cite{Shalit2017}: TARNet is a commonly used deep learning uplift model. Compared with X-learner (NN), it omits the additional imputed treatment effects fitting sub-models but introduces the shared layers for treated and control response networks. The shared network parameters could help alleviate the \textit{sample imbalance}.
	\item \textbf{CFR}\cite{Shalit2017}: CFR applies an additional loss to TARNet, which forces the learned treated and control covariate distributions to be closer. This could help to learn a balanced representation between those two groups and address the imbalance issue. We report the CFR performance using two distribution distance measurement loss functions, Wasserstein \cite{villani2008optimal,cuturi2014fast} (denoted as \textbf{CFR$_{wass}$}) and Maximum Mean Discrepancy (MMD) \cite{gretton2012mmd} (denoted as \textbf{CFR$_{mmd}$}).
	\item \textbf{X-network}: X-network is introduced in section \ref{Method Xnetwork}. It can be regarded as a variant of our final model DESCN where the ESN part is removed.
\end{itemize}

\subsection{Evaluation Metrics}
We adopt different commonly used uplift modeling evaluation metrics for different datasets based on whether the dataset contains the true uplift treatment effect value. More specifically, for \textit{Epilesy} synthetic dataset where the response generating process is known, we use the expected Precision in Estimation of Heterogeneous Effect $\epsilon_{PEHE}$ \cite{Schwab2019b, Shalit2017} and the absolute error in average treatment effect $\epsilon_{ATE}$:
\begin{align}
    \label{Pehe Ate}
    \begin{aligned}
    \epsilon_{PEHE} &= \frac{1}{n}\sum_{i} \big[\big(\hat\mu_1(x_i) - \hat\mu_0(x_i)\big)- \tau(x_i)\big]^{2} \\
    \epsilon_{ATE} &= \big|  \frac{1}{n}\sum_i \big(\hat\mu_1(x_i) - \hat\mu_0(x_i)\big) -  \frac{1}{n}\sum_i\tau(x_i) \big| \\
    \end{aligned}
\end{align}

Generally, $\epsilon_{PEHE}$ could better measure the CATE prediction accuracy at an individual level, while $\epsilon_{ATE}$ is a better indicator of the average effect difference across the given sample groups.

Since the ground truth for uplift value is impossible to retrieve on the production dataset, we report the normalized Area Under the Uplift Curve\footnote{Also known as \textit{Qini coefficient} in some literature.} (AUUC) \cite{Soltys2015-be, Zhao2017-kg, Rzepakowski2012-br, Gutierrez2016-co} value on the testing dataset, which is an indicator evaluating the uplift score ranking performance. We also compute the error of the Average Treatment Effect on the Treated group $\epsilon_{ATT}$ on this randomized test set:

\begin{align}
    \label{Att}
    \begin{aligned}
     ATT &= \frac{1}{|T|}\sum_{i\in T}{y_{i}} - \frac{1}{|C|}\sum_{i \in C}{y_{i}} \\ 
    \epsilon_{ATT} &= \big|\frac{1}{|T|} \sum_{i \in{T}} \big(\hat\mu_1(x_i) - \hat\mu_0(x_i)\big) - ATT\big| \\ 
    \end{aligned}
\end{align}

Besides the evaluation metrics, each experiment is repeated 5 times. Mean and standard error of all metrics are reported. Finally, we also calculate the relative improvement of a model over the baseline model for any evaluation metric $E$ as:
\begin{align}
    \label{impr}
    \begin{aligned}
    \textit{Impr(BaselineModel)} &= \frac{E(Model) - E(BaselineModel)}{E(BaselineModel)} \times 100\%
    \end{aligned}
\end{align}

\subsection{Hyper-parameter Settings}
We keep the same hyper-parameters in the common components across different models. Specifically, in the \textit{Epilepsy} dataset, all Neural Networks across different deep models consist of 128 hidden units with 3  fully connected layers. L2 regularization is applied to alleviate over-fitting with coefficient of 0.01, and the learning rate is set as 0.001 without decay. All the models are trained with a batch size of 500 and epoch as 15, the only different hyper-parameters are those associated with each specific network setting, such as the weights of different loss functions. Similarly, for the production dataset, all Neural Networks also consist of 3 fully connected layers. The shared network contains 128 hidden units and other sub-models contain 64 hidden units. The L2 regularization value is set as 0.001, and the learning rate is set as 0.001 with the decay rate as 0.95. All the models are trained with a batch size of 5000 and 5 epoches\footnote{More details regards experiment hyper-parameter settings could be found online.}.

\section{RESULTS AND DISCUSSIONS}
In this section, we present the overall performance of DESCN and the other methods to be compared. The following questions are also proposed and investigated:
\begin{itemize}
	\item Q1: Is the proposed Entire Space Network (ESN) useful in improving the overall performance and alleviating the \textit{treatment bias} issue? 
	\item Q2: Are the proposed cross network (X-network) and the pseudo treatment effect (PTE) network helpful in learning balanced treated and control response functions?
\end{itemize}
The experimental results of DESCN and other baseline models are summarized in Table \ref{table:main}.

\subsection{Overall Performance}
Based on the results from both datasets in Table \ref{table:main}, we have the following observations:
\begin{itemize}
    \item DESCN consistently outperforms the other baselines in both datasets in terms of different metrics, with reliability indicated by the small standard errors. Compared to CFRmmd, one of the most commonly used deep uplift models, DESCN achieves over +7\% relative improvement in $\epsilon_{PEHE}$ on the Epilepsy dataset, and +4\% relative improvement in AUUC on the highly biased production dataset. This indicates that DESCN is capable of handling both \textit{treatment bias} and \textit{sample imbalance} issues well and can achieve a higher performance in both individual-level treatment effect estimation accuracy and uplift ranking.
    \item DESCN achieves a more significant improvement in terms of $\epsilon_{ATE}$ and $\epsilon_{ATT}$ on the Epilepsy and the production dataset by +80\% and +84\%, respectively. With the help of X-network modeling treated and control distribution in a more balanced manner, DESCN could also improve the average treatment effect estimation accuracy for treated and control groups.
    
\end{itemize}

\subsection{Entire Space Network (ESN) Analysis (Q1)}
\label{ESN Ablation}
To further demonstrate the effectiveness by introducing the Entire Space Network (ESN) for jointly learning the treatment and response function in the entire sample space, we remove the ESN network from DESCN, which is the same as X-network as shown in Figure \ref{fig:Xnetwork}. The comparison of the performance with the full structure is as listed in the last two rows of Table \ref{table:main}:
\begin{itemize}
    \item From the table, we could find that after introducing ESN structure, DESCN improves X-network on $\epsilon_{PEHE}$ +3.7\% and $\epsilon_{ATE}$ +76.3\% on Epilepsy dataset. It also improves AUUC +4.9\% and $\epsilon_{ATT}$ +18.7\% on the production dataset. This indicates that by combining the learned treatment propensity score with learned responses in their individual sample space and directly learning ESTR and ESCR, the model could better capture the joint distribution in the entire sample space.
\end{itemize}
Moreover, the ESN is not limited to the DESCN model but also has the ability to extend to other existing individual response estimation methods to enhance the performance. We apply ESN to both TARNet and CFR in the same way as described in Section \ref{Method ESN}, and show their performance in Table \ref{tab:ESN result} and Table \ref{tab:ESN result 2} on both datasets:

\begin{table}[h]
    \centering
    \begin{tabular}{cccc}
        \toprule
        Model & $\sqrt{\epsilon_{PEHE}}$ &  & $\epsilon_{ATE}$\\
         & mean $\pm$ s.e. &  & mean $\pm$ s.e.\\
        \midrule
        TARNet & 0.1373 $\pm$ 0.0028 &  & 0.0405 $\pm$ 0.0091\\
        ESN+TARNet & 0.1320 $\pm$ 0.0008 & +3.9\% & 0.0233 $\pm$ 0.0053\\
        \midrule
        $CFR_{wass}$ & 0.1344 $\pm$ 0.0027 &  & 0.0305 $\pm$ 0.0069\\
        ESN+CFR$_{wass}$ & 0.1361 $\pm$ 0.0034 & -1.3\% & 0.0271 $\pm$ 0.0053\\
        \midrule
        CFR$_{mmd}$ & 0.1363 $\pm$ 0.0031 &  & 0.0263 $\pm$ 0.0097\\
        ESN+CFR$_{mmd}$ & 0.1577 $\pm$ 0.0013 & -15.7\% & 0.0183 $\pm$ 0.0063\\
        \bottomrule
    \end{tabular}
    \caption{Experimental results of applying ESN to TARNet and CFR on Epilepsy Dataset. Evaluated by $\epsilon_{PEHE}$ and $\epsilon_{ATE}$, the $\epsilon_{PEHE}$ improvement after applying ESN is included.}
    \label{tab:ESN result}
    \vspace{-0.5cm}
\end{table}

\begin{table}[h]
    \centering
    \begin{tabular}{cccc}
        \toprule
        Model & AUUC &  & $\epsilon_{ATT}$\\
         & mean $\pm$ s.e. &  & mean $\pm$ s.e.\\
        \midrule
        TARNet & 0.0309 $\pm$ 0.0021 &  & 0.0106 $\pm$ 0.0016\\
        ESN+TARNet & 0.0340 $\pm$ 0.0008 & +10.0\% & 0.0165 $\pm$ 0.0017\\
        \midrule
        $CFR_{wass}$ & 0.0261 $\pm$ 0.0002 &  & 0.0266 $\pm$ 0.0013\\
        ESN+CFR$_{wass}$ & 0.0264 $\pm$ 0.0018 & +1.1\% & 0.0212 $\pm$ 0.0015\\
        \midrule
        CFR$_{mmd}$ & 0.0324 $\pm$ 0.0029 &  & 0.0258 $\pm$ 0.0015\\
        ESN+CFR$_{mmd}$ & 0.0331 $\pm$ 0.0005 & +2.1\% & 0.0207 $\pm$ 0.0010\\
        \bottomrule
    \end{tabular}
    \caption{Experimental results of applying ESN to TARNet and CFR on Production Dataset. Evaluated by AUUC and $\epsilon_{ATT}$, the AUUC improvement after applying ESN is included.}
    \label{tab:ESN result 2}
    \vspace{-0.5cm}
\end{table}
\begin{itemize}
    \item With the help of alleviating treatment bias issue by learning treatment and response functions in the entire sample space, ESN enhances the performance of TARNet by +3.9\% in $\epsilon_{PEHE}$ on Epilepsy Dataset and 10.0\% in AUUC on the production dataset. This indicates that ESN is capable of enhancing the individual ITE estimation accuracy and uplifting ranking performance. However, it does not show much improvement over CFR on the Epilepsy dataset and with a slight improvement on the production dataset over those two metrics. A possible reason is that CFR already uses the additional loss function (either WASS or MMD) to enforce a similar representation of treated and control response in their individual sample space. It might be in conflict with the additional propensity information learned in the ESN.
    \item Furthermore, we observe that ESN improves $\epsilon_{ATE}$ +42.5\% on TARNet, +11.1\% on CFRwass, and +30.4\% on CFRmmd on the Epilepsy dataset. For the production dataset, although it does not improve on $\epsilon_{ATT}$ on the TARNet, it largely improves $\epsilon_{ATT}$ on CFRwass and CFRmmd by +20.3\% and +19.8\% respectively. A similar trade-off between estimating average effect and estimating individual effect is also observed in \cite{Shalit2017}. 
    \item We believe that debiasing treated and control distribution through ESN or loss functions in CFR will improve the individual estimation accuracy but hurt the average estimation performance to a certain degree. In the production dataset, the treatment bias is more significant than it in the Epilepsy dataset, hence the ESN+TARNet performs slightly worse than TARNet in $\epsilon_{ATT}$. Nonetheless, ESN could still slightly improve CFR performance on both datasets, which shows that training ESTR and ESCR in the entire sample space is still beneficial.
    \item  We can explore why the ESN module in DECSN can reduce the treatment bias from the ATE perspective. Firstly, according to the IPW(inverse probability 
 weighting)\cite{rosenbaum1983central} theory we know that  $ATE= \mathbb{E}_\mathbb{P} \{\frac{W_i \cdot Y_i}{\pi(X_i)}\}- \mathbb{E}_\mathbb{P}\{\frac{(1-W_i)\cdot  Y_i}{1-\pi(X_i)}\} $  
. According to Eq.(3) in section 3.2, we can expand the equation as: 

      $ \begin{aligned} ATE & = \frac{1}{N}  \sum_{i}  \frac{ \mathbb{E}_\mathbb{P} \{W_i\cdot Y_i|X_i\} }{\pi(X_i)} - \frac{1}{N}  \sum_{i} \frac{  \mathbb{E}_\mathbb{P}\{(1-W_i)\cdot Y_i|X_i\}} {1-\pi(X_i)} \\ 
      & = \frac{1}{N}  \sum_{i}   \frac{ ESTR_i }{\pi(X_i)} - \frac{1}{N}  \sum_{i}  \frac{  ESCR_i }{ 1- \pi(X_i)}  \\
& = \frac{1}{N}  \sum_{i}   \mu_1(X_i) - \frac{1}{N}  \sum_{i}   \mu_0(X_i) \end{aligned}  $ 
At this point, we can see its mechanism to reduce the treatment bias.

\end{itemize}

\subsection{Cross Network (X-network) Analysis (Q2)}
Recall that the design of the cross network (X-network) utilizes a specially designed intermediate Pseudo Treatment Effect (PTE) variable to connect with both Treat Response (TR) and Control Response (CR) to balance the learned response functions. As the X-network structure only differs with TARNet in this additional PTE Network, we compare the performance of X-network and TARNet in Table \ref{table:main}. 

We can see that X-network improves the $\epsilon_{PEHE}$ +9.6\% on the Epilepsy dataset and AUUC +10.0\% on the production dataset. Furthermore, there also exists significant improvement in $\epsilon_{ATE}$ and $\epsilon_{ATT}$. All these prove that the designed X-network is capable of learning each response function more balanced, and can well simulate the hidden treatment effect. 





\section{Conclusion and Future Work}
In this paper, we propose an integrated multi-task model for Individual Treatment Effect (ITE) estimation, Deep Entire Space Cross Networks (DESCN). It introduces an Entire Space Network (ESN) to jointly learn the distribution of treatment and response functions in the entire sample space to address the \textit{treatment bias} issue. Moreover, a novel cross network (X-network) is presented which is capable of directly modeling hidden treatment effects, and also has the advantage of alleviating distribution \textit{sample imbalance}. Finally, DESCN combines ESN and X-network in a multi-task learning manner, which integrates information of the treatment propensity, the response, and the hidden treatment effect through a cross network. Extensive experiments in both synthetic dataset adapted from the ACIC data challenge and a large-scale voucher distribution production dataset are conducted. The results show that DESCN achieves over +4\% to +7\% improvement in $\epsilon_{PEHE}$ and AUUC on the synthetic dataset and production dataset respectively. This demonstrates that DESCN is effective in ITE estimation accuracy and uplift ranking.

There are several potential future works for exploration. In industrial applications, multiple treatments could be given at the same time. An extension of DESCN to the multi-treatment setting could be studied by extending the X-network to multiple response functions. Another direction is to extend DESCN to fit continuous treatment outcomes, which also has wide applications.


\bibliographystyle{ACM-Reference-Format}
\bibliography{reference}


\end{document}